\documentclass[10pt,twocolumn,letterpaper]{article}

\usepackage{cvpr}              
\usepackage{algorithmic}
\usepackage{algorithm}
\usepackage{multirow}
\usepackage{amsmath}
\usepackage{comment}
\usepackage{xcolor}
\usepackage{colortbl}
\usepackage{CJKutf8}

\newcommand{\myparagraph}[1]{%
  \vspace{1mm}
  \noindent\textbf{#1}%
}
\usepackage[accsupp]{axessibility} 

\definecolor{cvprblue}{rgb}{0.21,0.49,0.74}
\usepackage[pagebackref,breaklinks,colorlinks,allcolors=cvprblue]{hyperref}

\def\paperID{17766} 
\def\confName{CVPR}
\def\confYear{2025}

\title{Generating 6DoF Object Manipulation Trajectories \\ from Action Description in Egocentric Vision}

\author{
Tomoya Yoshida$^{1}$ \;\; Shuhei Kurita$^{2}$ \;\; Taichi Nishimura$^{3}$ \;\; Shinsuke Mori$^{1}$
\\
$^1$Kyoto University \;\;
$^2$National Institute of Informatics \;\;
$^3$Sony Interactive Entertainment \\
}

\usepackage{xspace}

\newcommand{\datasetname}{EgoTraj\xspace}
\def\Bdma#1{\mbox{\boldmath{$#1$}}}

\usepackage[font=small]{caption}

\begin{document}

\makeatletter
\let\@oldmaketitle\@maketitle
\renewcommand{\@maketitle}{\@oldmaketitle
\vspace*{-5mm}
\includegraphics[width=\textwidth]{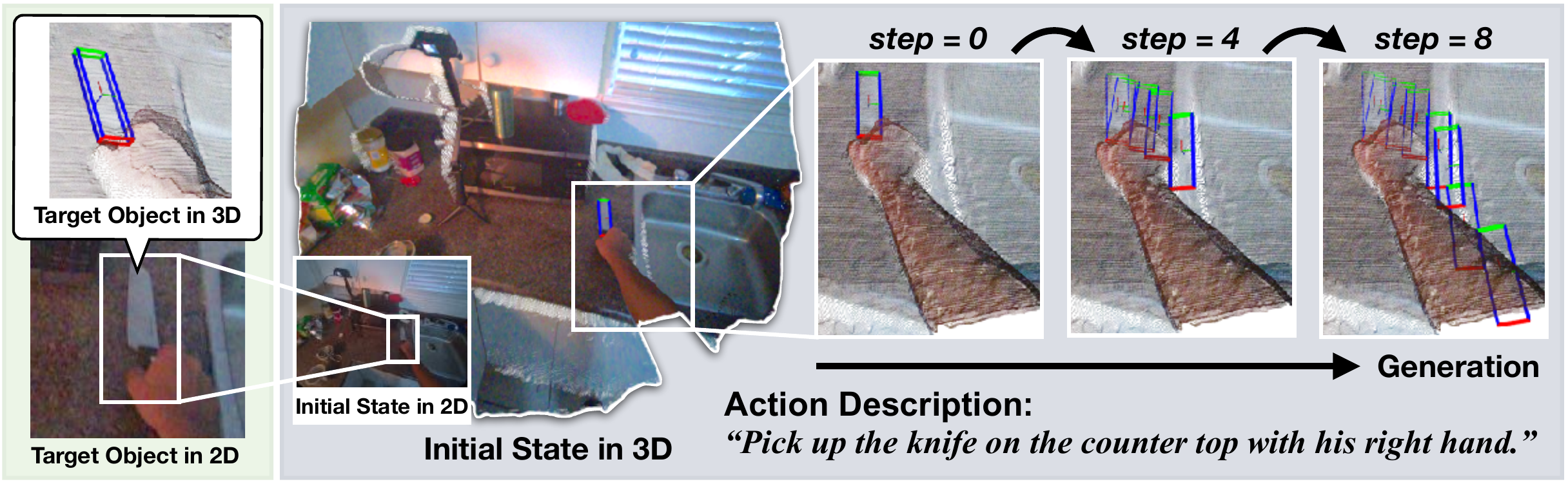}
  \centering
  \vspace*{-7mm}
  \captionof{figure}{\small \textbf{6DoF object manipulation trajectory}. This task aims to generate a sequence of 6DoF object poses from an action description and an initial state comprising the visual input and the object's initial pose.}
  \vspace*{-0mm}
  \label{fig:overview}
  \bigskip}
\makeatother

\maketitle

\begin{abstract}

\vspace{-3mm}

\makeatletter
\newcommand\blfootnote[1]{%
  \begingroup
  \renewcommand\thefootnote{}%
  \footnote{#1}%
  \addtocounter{footnote}{-1}%
  \endgroup
}
\makeatother

\blfootnote{This work is also done in Research and Development Center for Large Language Models, National Institute of Informatics (NII LLMC).}

Learning to use tools or objects in common scenes, particularly handling them in various ways as instructed, is a key challenge for developing interactive robots. Training models to generate such manipulation trajectories requires a large and diverse collection of detailed manipulation demonstrations for various objects, which is nearly unfeasible to gather at scale. In this paper, we propose a framework that leverages large-scale ego- and exo-centric video datasets --- constructed globally with substantial effort --- of Exo-Ego4D to extract diverse manipulation trajectories at scale. From these extracted trajectories with the associated textual action description, we develop trajectory generation models based on visual and point cloud-based language models. 
In the recently proposed egocentric vision-based in-a-quality trajectory dataset of HOT3D, we confirmed that our models successfully generate valid object trajectories, establishing a training dataset and baseline models for the novel task of generating 6DoF manipulation trajectories from action descriptions in egocentric vision. 
Our dataset and code: \url{https://biscue5.github.io/egoscaler-project-page/}.

\end{abstract}
    
\section{Introduction}
\label{sec:intro}

Use of tools and manipulation of objects at their will is an essential ability for developing robots that assist human activities. Especially having knowledge of manipulating numerous tools or objects in our work places and houses following the requests from humans is the next key technology for developing interactive robots that support us in daily scenes. One way to achieve this is to preliminarily learn various ways of manipulating objects from the human demonstration, as the recent robotic studies suggest that human demonstration acts as quite effective supervision for object manipulation trajectory generation~\cite{rt-x, aloha, mobile-aloha}. However, this human demonstration-driven approach evokes the well-known limitation of data scarcity: the object manipulation trajectories from real human demonstrations are quite informative yet extremely limited and almost prohibitive for covering numerous objects we use in our daily lives.

One promising solution for this is to make use of videos of daily working scenes for extracting human demonstrations of various object manipulations. Egocentric videos are becoming popular in virtual and augmented reality (VR/AR)~\cite{dehghan2021arkitscenes} and even recording our daily life activities~\cite{aria-glasses, quest3} in both industry and research fields. For tracking human motions, HOT3D~\cite{hot3d} provides quality 3D poses and models of hands and objects from precise lab recordings of human activity. Unfortunately, the precise recording of human motion trajectories is still considerably costly, and hence such data is still limited for developing models that generate object manipulation trajectories whilst general purpose egocentric recordings are becoming available at scale as of the Ego4D~\cite{ego4d} and Ego-Exo4D~\cite{egoexo4d} datasets. Hereby, we assume extracting human motions for manipulating objects from egocentric vision datasets at scale is the key break-through for developing interactive robots.

In this paper, we introduce a framework to extract an object manipulation trajectory, a sequence of 6DoF object positions and rotations, from egocentric videos.
Our framework consists of the following: determination of temporal action span, extracting objects with the open-vocabulary segmentation model~\cite{grounded-sam} and tracking them with the dense 3D point tracker~\cite{spatracker}, determining the camera coordinates, and obtaining the object rotation by the singular value decomposition from the translation between scenes.
Applying this framework to egocentric videos from Ego-Exo4D~\cite{egoexo4d}, we construct \datasetname with 28,497 object manipulation trajectories and corresponding action descriptions.
\cref{tab:statistics} presents the dataset statistics from previous studies and ours. Our dataset emerges as the largest, containing 28,497 action descriptions and trajectories. While existing datasets such as H2O~\cite{h2o}, EgoPAT3D~\cite{egopat3d}, and HOT3D~\cite{hot3d} rely on lab-recordings with multiple cameras or depth cameras, our dataset is extracted from ordinary egocentric videos with a single camera without pre-recorded camera coordinates or depth, and encompasses an entirely different order of magnitude in the variety of action verbs and objects from real-world scenarios. As our method does not rely on pre-recorded camera coordinates, our method is applicable for most of egocentric video datasets, covering diverse object manipulations from egovision.

With the extracted trajectories of \datasetname, we develop a branch of models to generate object manipulations from the textual action description. The overview of the task, including extracted 6DoF object trajectory, is presented in \cref{fig:overview}.
In experiments, we develop object manipulation trajectory generation models based on visual and point cloud-based language models with discretized trajectory tokens~\cite{rt-1, rt-2, rt-x, openvla}.
Our experimental results on the HOT3D dataset demonstrate that our models successfully generate valid object manipulation trajectories.
Furthermore, we conduct the experiments of generating action descriptions using our object trajectories, suggesting that utilizing trajectories enhances 
the generation performance, particularly for the similarity of verbs in the action descriptions. 
To the best of our knowledge, this is the first attempt to extract large-scale 6DoF object manipulation trajectories from egocentric videos without pre-recored camera coordinates and develop visual and point cloud-based language models for generating object manipulation trajectories from textual action descriptions.
\section{Related Work}
\subsection{Object Interaction Understanding}
Understanding object interaction involves two key factors of object affordance and manipulation motion.
Koppula and Saxena~\cite{anticipate-affordance} addressed human action anticipation by using object affordances to predict future motion trajectories. They demonstrated that understanding object affordance and predicting motion trajectories is beneficial for reactive robotic response. Subsequently, this field has been extensively explored in the contexts of object affordance learning~\cite{affordance-net} and object interaction prediction~\cite{ego4d}. 

\myparagraph{Object Affordance Learning.} 
Object affordance learning aims to localize objects~\cite{learning-to-act, phrase-based-aff, tdod} or their specific parts~\cite{handal, affgrasp, madiff, open-vocab-aff, locate} necessary for performing certain actions. 
This task has gained attention for applications in robotic manipulation~\cite{vrb, voxposer, rt-affordance}.
However, most studies focus on grasping affordance, limiting various manipulation understandings.
More recently, hand-object interaction generation~\cite{reconstruct-object, contact-db, gan-hand, text2hoi, hoi-diffusion} has been proposed, addressing both affordance learning and manipulation understanding in 3D space. This task builds upon richly annotated video datasets~\cite{hoi4d, h2o} that include hand poses, object meshes, and scene point clouds.

\begin{table}[t]
\centering
\footnotesize
{
\footnotesize
\begin{tabular}{lrrrrr}
\toprule
\multicolumn{1}{c}{Dataset} & \multicolumn{1}{c}{\#AD} & \multicolumn{1}{c}{\#Verb} & \multicolumn{1}{c}{\#Object} & \multicolumn{1}{c}{\#Trajectories} \\
\midrule
EgoPAT3D$^\dagger$~\cite{egopat3d} & - & 26 & 11 & 11,141 \\
H2O$^\dagger$~\cite{h2o} & - & 12 & 8 & 13,653 \\
\midrule
HOT3D~\cite{hot3d} & 986 & 30 & 33 & 986 \\
\datasetname (Ours) & \textbf{28,497} & \textbf{228} & \textbf{4,158} & \textbf{28,497} \\
\bottomrule
\end{tabular}
\vspace{-3mm}
}
\caption{Dataset statistics from previous studies and ours. The ``\#AD'' column denotes the number of action descriptions. ``\#Trajectories'' of $\dagger$ is repored in \cite{3d-hand-traj}. Note that our dataset is constructed through a machine learning-based approach, while others are through precise annotations.
}
\vspace{-3mm}
\label{tab:statistics}
\end{table}

\myparagraph{Object Interaction Prediction.}
Object interaction prediction includes various tasks such as interaction anticipation~\cite{forecast-hoi, Furnari_2019_ICCV, forecast-hoi, Miech_2019_CVPR_Workshops, technical-antipation}, next active object detection~\cite{anticipate-affordance, forecast-hoi, Yang_2024_CVPR, Thakur_2024_WACV, active-objdet, naod-ego, ego-aod}, and hand trajectory forecasting~\cite{hoi-forecast, madiff, diff-ip2d, pear}. These tasks have potential applications such as VR/AR~\cite{so-predictable}, human-robot interaction~\cite{aided-htf, egopat3d-v2}.
However, most studies addressed these tasks in 2D space, limiting their applications.
To mitigate this limitation, recent work has extended these tasks into 3D space~\cite{3d-hand-traj, egopat3d}. For instance, Bao \etal~\cite{3d-hand-traj} addressed 3D hand trajectory forecasting by proposing an uncertainty-aware state space transformer, extending conventional 2D trajectory forecasting task to 3D.
Despite these advancements, addressing diverse manipulation motions remains challenging due to the high cost of annotations in 3D space. In contrast, our work proposes a framework to extract 6DoF object manipulation trajectories from egocentric videos, enabling the handling of diverse manipulation motions. From these extracted trajectories with the associated action descriptions, we address text-guided object manipulation trajectory generation.

\begin{figure*}[t]
  \centering
  \includegraphics[width=\textwidth]{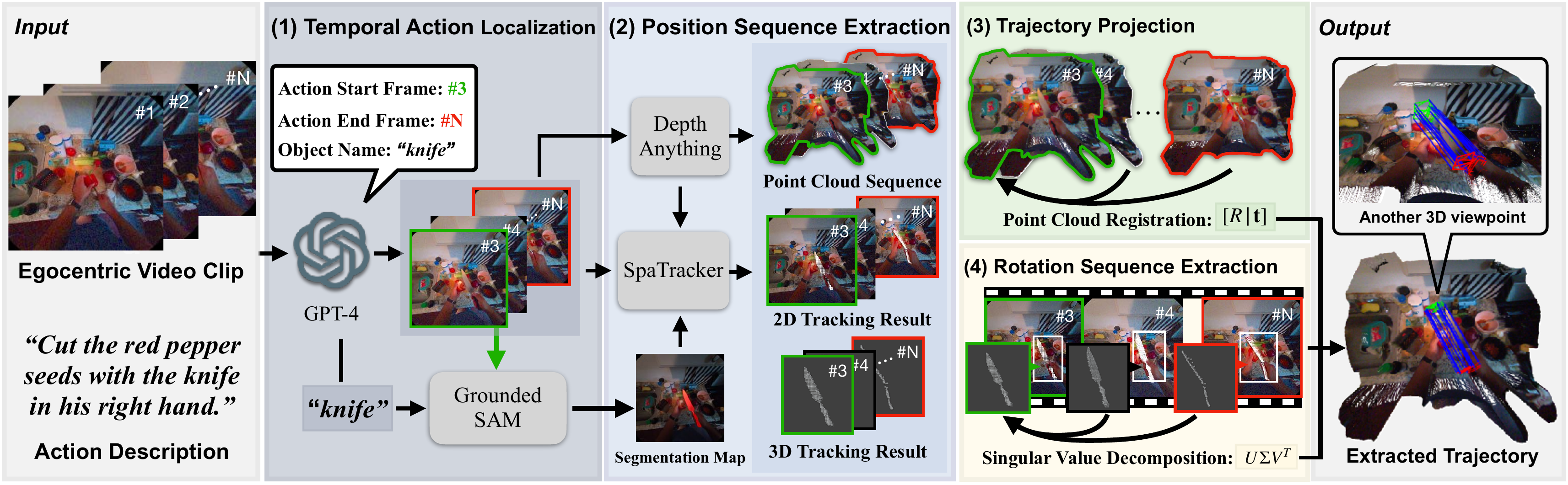}
  \vspace{-3mm}
  \caption{\textbf{Trajectory extraction from egocentric videos.}
  Four steps of (1) temporal action localization, (2) position sequence extraction, (3) trajectory projection, and (4) rotation sequence extraction.
  Our project page includes visualization of resulting trajectories with videos.}
  \vspace{-3mm}
  \label{fig:method}
\end{figure*}

\subsection{Egocentric Vision Dataset}
Egocentric videos encompass diverse object manipulation and hand activities, serving as potential resources for object interaction learning.
A wide range of datasets and benchmarks have been proposed~\cite{ego4d, egoexo4d, epic, rescaling, assembly101, egtea}. For example, Ego-Exo4D~\cite{egoexo4d} is a large-scale egocentric-exocentric dataset, providing rich multi-modal resources including gaze, point clouds, and texts. 
Based on these large-scale datasets, multiple detailed tasks have been developed~\cite{ego3dt, out-of-sight, goalstep, epic-field, hoi-forecast, Rodin_2024_CVPR}. Additionally, several datasets have also been developed for specific purposes, including hand-object-interaction and VR tasks~\cite{hot3d, holo-assist, h2o, hoi4d, honnotate}.
HOT3D~\cite{hot3d} is an egocentric dataset for 3D hand and object tracking. 
Trajectories are captured using optical markers and OptiTrack cameras, resulting in precise hand poses and 6DoF object trajectories.
While several other datasets~\cite{hoi4d, honnotate, h2o} also provide similar hand and object poses, HOT3D covers a wider range of trajectories involving diverse objects and action scenarios.
To this end, we utilize 6DoF object manipulation trajectories from HOT3D to validate model performance.

\subsection{Object Pose Estimation / Tracking}
Estimating object pose from visual sensors, termed object pose estimation, is a crucial topic for robotic perception~\cite{pose-review}. Object pose tracking is an extended task of object pose estimation, aiming to efficiently track temporal object poses~\cite{deepim, foundationpose}. Methods in both tasks often rely on objects' CAD models~\cite{pvn3d, pix2pose, deepim} or several reference images~\cite{onepose, onepose++, fs6d, foundationpose}. 
Although pose estimation from visual input is similar to our task, we aim to understand and generate pose transformations during manipulation.
Consequently, our task does not require any 3D object models or reference images of objects.

\subsection{Language Models for Visual / Robotic Tasks}
Recent advances in large auto-regressive models, such as large language models (LLMs) and vision-language models (VLMs), have established these models as powerful tools in computer vision~\cite{minigpt-3d, blip2, video-llama, llava, llava1.5, 3d-vista, ulip2, 3d-llm} and robotics~\cite{rt-1, rt-2, openvla}. For instance, PointLLM~\cite{pointllm}, a 3D-LLM for object point cloud understanding, has shown strong performance in 3D object captioning and has also demonstrated capability in scene-level point cloud understanding.
Moreover, by embedding spatio-temporal information into language space, several studies have achieved comparable or superior performance to conventional approaches in image, video, and point cloud understanding benchmarks, such as object detection~\cite{ofa, pix2seq}, object tracking~\cite{omnivid}, temporal action localization~\cite{lita}, and 2D/3D visual grounding~\cite{ofa, 3d-vista}. Inspired by these studies, we develop trajectory generation models based on VLMs, formalizing our task as next token prediction task.

\section{Dataset Construction Framework}

We construct a large-scale training dataset containing 6DoF object manipulation trajectories along with images, depth maps, and action descriptions from an egocentric vision dataset of Ego-Exo4D~\cite{egoexo4d}. Due to the cost and scalability, we propose an automated approach here.
Note that our framework does not require any camera extrinsic parameters in advance.
Additionally, we utilize precise 6DoF object manipulation trajectories extracted from HOT3D~\cite{hot3d} dataset for the evaluation set.

\subsection{Training Data Creation}
\label{sec:training-data}
As shown in \cref{fig:method}, our pipeline consists of four stages. First, given an egocentric video, we determine the start and end timestamps of the action and identify the manipulated object within the scene. Second, we extract the position sequence of the manipulated object using an open-vocabulary segmentation model~\cite{grounded-sam} and a dense 3D point tracker~\cite{spatracker}. Third, we project the sequence into the camera coordinate system of the first frame using point cloud registration. Fourth, we extract a rotation sequence by computing the transformation between the two object point clouds using singular value decomposition.

\myparagraph{Temporal Action Localization.}
Based on the original annotated timestamps, we first split the raw egocentric video into multiple clips. Each clip spans a four-second interval centered on each timestamp.
Next, we sample image frames within each clip and assign them sequential indices. We input these indexed frames along with the corresponding action description, originally annotated in Ego-Exo4D, into OpenAI GPT-4o~\cite{gpt-4} to determine the action's start and end timestamps: $t_{\text{start}}$ and $t_{\text{end}}$. We denote the frames at $t_{\text{start}}$ and $t_{\text{end}}$ as $f_{\text{start}}$ and $f_{\text{end}}$, respectively.
Moreover, we use the model to extract the manipulated object name and determine whether the object is rigid or not.
We filter out non-rigid objects as the transformation of the non-rigid objects is left for future work. We provide the system prompt in the Appendix.

\myparagraph{Position Sequence Extraction.}
We first use an open-vocabulary object segmentation model~\cite{grounded-sam} to identify the manipulated object in the first frame $f_{\text{start}}$. This model takes $f_{\text{start}}$ and the object name as inputs, and outputs a 2D segmentation map. Next, we employ an off-the-shelf monocular depth estimation model~\cite{depth-anything} and SpaTracker~\cite{spatracker}. SpaTracker is a state-of-the-art dense point tracking model that tracks any 2D points in 3D space.
Based on a sequence of frames between $t_{\text{start}}$ and $t_{\text{end}}$, we obtain depth maps for each frame using a depth estimation model. 
For SpaTracker, we input the sequence of frames, the corresponding depth maps, and the segmentation map into the model. The model then outputs 3D tracking results of the segmented object.

\myparagraph{Trajectory Projection.}
Motions in egocentric videos can be separated into the object motions and the camera motions by the camera wearer.
Due to the camera motions, tracking results are not aligned with the camera coordinate system of the start frame $f_{\text{start}}$. However, the camera motion between two consecutive frames can be assumed to be small and is represented by a projection matrix. 
To obtain the projection matrix between two consecutive frames, we use Depth Anything~\cite{depth-anything} to estimate depth, convert each RGB-D image frame into point clouds, and then apply point cloud registration to determine the projection matrix of the camera extrinsic parameters.
We successfully obtain the sequence of the camera extrinsic parameters from each frame to $f_{\text{start}}$ by multiplying these projection matrices.  
Recently, DUSt3R~\cite{dust3r_cvpr24} can jointly perform camera pose estimation and point cloud construction, internally using point cloud registration of RoMa~\cite{bregier2021deepregression} and gradient-based optimization. However, we notice DUSt3R can squeeze detailed 3D motions of objects or hands in egocentric views, leaving this direction for future work.

Each 2D image frame is first converted into a point cloud using an estimated depth map. Let $(i, j)$ denote the image pixel coordinates and $d_{ij}$ the corresponding depth, then
\begin{equation}
    \begin{bmatrix}
        x \\
        y \\
        z
    \end{bmatrix}
    = d_{ij} K^{-1} 
    \begin{bmatrix}
        i \\
        j \\
        1
    \end{bmatrix},
\end{equation}
where $K$ is the camera's intrinsic parameters. Next, we compute the normals and Fast Point Feature Histograms (FPFH)~\cite{fpfh} for each point cloud to perform registration. For initial alignment, we use RANSAC-based~\cite{ransac} feature matching between two point clouds using these FPFH features. We further refine the alignment using the colored iterative closest point algorithm~\cite{color-pcr}. 
With the above operations, we obtain camera extrinsic parameters for each frame. The 3D tracking results are then projected to the camera coordinate system of $f_{\text{start}}$ by multiplying the coordinates at each timestep by the corresponding projection matrix.

\myparagraph{Rotation Sequence Extraction.}
As shown in \cref{fig:method}, SpaTracker effectively handles self-occlusion, enabling the computation of temporal object pose transformations. Given the tracking result of the manipulated object, the object pose is computed using singular value decomposition (SVD). 
First, we set the initial pose of the manipulated object to an identity transformation: $[0, 0, 0]^{T}$. We then compute the covariance matrix $H$ between the object point cloud at the initial timestamp and each subsequent timestamp. Applying singular value decomposition to $H$ gives $H = U \Sigma V^{T}$. Finally, we compute the rotation matrix $R = V U^{T}$, where $R \in \mathbb{R}^{3 \times 3}$, and convert $R$ into a rotation vector $r \in \mathbb{R}^{3}$. By concatenating the position and rotation sequences, we obtain the 6DoF object manipulation trajectory. Additionally, we calculate the minimum 3D bounding box from the object point cloud at the initial timestamp.

\subsection{Evaluation Data}
\label{sec:eval-data}
For evaluation, we utilize the existing 6DoF hand and object tracking dataset of HOT3D~\cite{hot3d}.
HOT3D is an egocentric view dataset for 3D hand and object tracking. The dataset is constructed using optical markers and multiple infrared OptiTrack cameras, yielding precise hand and object 6DoF information. The dataset is recorded with Project Aria glasses~\cite{aria-glasses} and Quest 3~\cite{quest3}.
Unfortunately, the dataset does not include temporal action descriptions, resulting in a misalignment with our task setting. To mitigate this difference, we split the raw videos and determine the action start and end timestamps with action descriptions as described in \cref{sec:training-data}. Additionally, we obtain depth maps using the depth estimation model and align the trajectories with the estimated depth maps for models that require depth or point clouds. Further details are provided in the Appendix.

\section{6DoF Object Manipulation Generation}
\myparagraph{Task Setting.}
Our task is to generate a sequence of 6DoF object poses for object manipulation, based on an action description and an initial state comprising a visual input and the initial pose of the object. Each pose in this sequence is defined by the position of the object's centroid and its rotation in 3D space.

\begin{figure}[t]
  \centering
  \includegraphics[width=\linewidth]{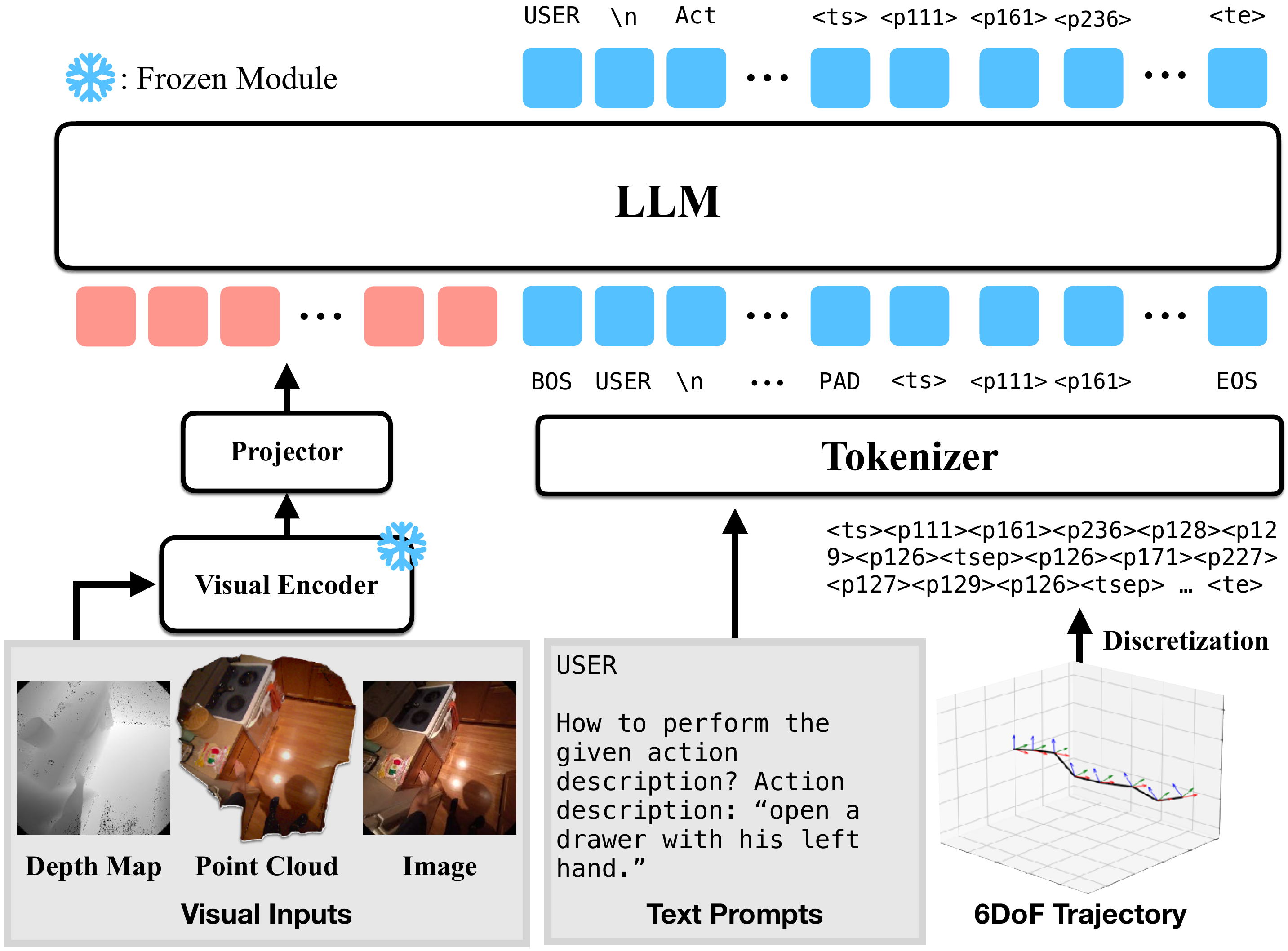}
  \vspace{-3mm}
  \caption{\textbf{Overview of model architecture.} Our model architecture utilizes visual and point cloud-based language models as backbones and extends them by incorporating extended vocabularies for trajectory tokenization.}
  \vspace{-3mm}  
  \label{fig:model}
\end{figure}

\myparagraph{Model.}
Considering recent advancements in multi-modal language models~\cite{omnivid, ofa, lita, pix2seq}, we develop object manipulation trajectory generation models based on visual and point cloud-based language models (VLMs), formalizing our task as next token prediction task.
This is achieved by incorporating an extended vocabulary for trajectory tokenization into the VLMs. 
An overview of the model architecture is depicted in Figure \ref{fig:model}.

\myparagraph{Discretization of object poses.}
Inspired by robotic foundation models~\cite{rt-1, rt-2, rt-x, openvla}, we discretize each continuous dimension into 256 bins.  For each variable, we map the target variable to one of the 256 bins, where the bins are uniformly distributed within the bounds of each variable. Using this discretization, we obtain an $N \times 6$ array of discrete integers from a trajectory, where each point $(x, y, z, \text{roll}, \text{pitch}, \text{yaw}) \in [0, 255]^6$. We reserve $N=256$ special tokens as trajectory tokens.

\myparagraph{Backbones.}
As our framework is compatible with any vision-language models (VLMs), we employ several VLMs as the backbone architecture.
\begin{itemize}
    \item \textbf{BLIP-2}: BLIP-2~\cite{blip2} is a VLM for image captioning and visual question answering. The model uses a pre-trained vision transformer from CLIP~\cite{clip} and OPT (2.7B/6.7B)~\cite{opt} as its language decoder.
    \item \textbf{VILA}: VILA~\cite{vila} is a state-of-the-art model for downstream vision-and-language tasks. It uses a similar architecture to BLIP-2 and is capable of reasoning over multiple images.
    \item \textbf{PointLLM}: PointLLM~\cite{pointllm} is a 3D-LLM designed for 3D vision tasks, such as 3D question answering and 3D grounding. The model uses Point-BERT~\cite{point-bert} from ULIP-2~\cite{ulip2} as a colored point cloud encoder and LLaMA~\cite{llama} with the Vicuna (7B/13B)~\cite{vicuna} checkpoints as the language decoder.
    \item \textbf{MiniGPT-3D}: MiniGPT-3D~\cite{minigpt-3d} is a light-weight 3D-LLM for downstream 3D vision understanding tasks. This model uses a similar architecture to PointLLM and incorporates an extended Mixture of Experts~\cite{moe, moe-em}.
\end{itemize}
As existing VLMs do not utilize depth information, we extend BLIP-2~\cite{blip2} and VILA~\cite{vila} by incorporating an additional encoder for depth maps. To evaluate the effectiveness of depth information, we use a pre-trained image encoder as the depth encoder for each model. We concatenate the encoded depth features with image features and provide them as inputs to LLM. For the objective function, we employ cross-entropy loss across all backbone models.
\begin{table*}[t]
\centering
{
\footnotesize
\begin{tabular}{lcrrrrrrrrr}
\toprule
 & & \multicolumn{4}{c}{\textit{3DoF Training}} & \multicolumn{5}{c}{\textit{6DoF Training}} \\
\cmidrule(rl){3-6}\cmidrule(rl){7-11}
 & & \multicolumn{2}{c}{\textit{3D pos.}} & \multicolumn{2}{c}{\textit{2D pos.}} & \multicolumn{2}{c}{\textit{3D pos.}} & \multicolumn{2}{c}{\textit{2D pos.}} &\multicolumn{1}{c}{\textit{3D rot.}} \\
\cmidrule(rl){3-4}\cmidrule(rl){5-6}\cmidrule(rl){7-8}\cmidrule(rl){9-10}\cmidrule(rl){11-11}
 Model & Input & \multicolumn{1}{c}{ADE} & \multicolumn{1}{c}{FDE } & \multicolumn{1}{c}{ADE} & \multicolumn{1}{c}{FDE} & \multicolumn{1}{c}{ADE} & \multicolumn{1}{c}{FDE } & \multicolumn{1}{c}{ADE} & \multicolumn{1}{c}{FDE} & \multicolumn{1}{c}{GD} \\
\midrule
 Seq2Seq~\cite{attention-is-all} & Image & \textbf{0.269} & 0.475 & \underline{0.215} & 0.372 & 0.374 & 0.625 & 0.325 & 0.528 & 0.559 \\
 BLIP-2 (2.7B)~\cite{blip2} & Image & 0.278 & \underline{0.464} & 0.218 & \underline{0.346} & 0.280 & \underline{0.465} & 0.218 & 0.344 & 0.545 \\
 BLIP-2 (6.7B)~\cite{blip2} & Image & 0.286 & 0.487 & 0.224 & 0.365 & 0.286 & 0.475 & 0.219 & 0.349 & 0.543 \\
 VILA (3B)~\cite{vila} & Image & 0.500 & 0.662 & 0.428 & 0.546 & 0.477 & 0.619 & 0.390 & 0.489 & 0.826 \\
 VILA (8B)~\cite{vila} & Image & 0.316 & 0.512 & 0.249 & 0.388 & 0.293 & 0.478 & 0.225 & 0.347 & 0.545 \\
\midrule
 Seq2Seq~\cite{attention-is-all} & Image + Depth & 0.302 & 0.541 & 0.250 & 0.438 & 0.341 & 0.558 & 0.280 & 0.452 & 0.590 \\
 BLIP-2 (2.7B)~\cite{blip2} & Image + Depth & 0.288 & 0.481 & 0.227 & 0.363 & \underline{0.275} & \textbf{0.458} & \underline{0.211} & \underline{0.332} & 0.543 \\
 BLIP-2 (6.7B)~\cite{blip2} & Image + Depth & 0.283 & 0.477 & 0.218 & 0.351 & 0.282 & 0.469 & 0.219 & 0.345 & \underline{0.542} \\
 VILA (3B)~\cite{vila} & Image + Depth & 0.301 & 0.492 & 0.229 & 0.355 & 0.294 & 0.480 & 0.223 & 0.344 & \textbf{0.541} \\
 VILA (8B)~\cite{vila} & Image + Depth & 0.298 & 0.494 & 0.234 & 0.368 & 0.318 & 0.513 & 0.253 & 0.391 & 0.563 \\
\midrule
 MiniGPT-3D (2.7B)~\cite{minigpt-3d} & Point cloud &  0.299 & 0.487 & 0.236 & 0.368 & 0.281 & 0.467 & 0.218 & 0.342 & 0.544  \\
 PointLLM (7B)~\cite{pointllm} & Point cloud & \underline{0.274} & \textbf{0.459} & \textbf{0.210} & \textbf{0.328} & \textbf{0.271} & \textbf{0.458} & \textbf{0.208} & \textbf{0.327} & \textbf{0.541} \\
\bottomrule

\end{tabular}
}
\vspace{-2mm}
\caption{
\textbf{Comparison of 3DoF and 6DoF object manipulation trajectory generation.} The ``Input'' column indicates the visual modality input for each model. Lower values are preferable for all metrics. Best and secondary results are viewed in \textbf{bold} and \underline{underline}.
}
\vspace{-4mm}
\label{tab:results}
\end{table*}

\section{Experiments}
\subsection{Experimental Setting}
\myparagraph{Dataset.}
We apply our framework (\cref{sec:training-data}) to Ego-Exo4D~\cite{egoexo4d} dataset, resulting in collecting 28,497 samples.
We split them into 27,788 training samples and 709 validation samples. Additionally, we use data extracted from HOT3D~\cite{hot3d} as the test split, resulting in 986 test samples. 
We sample image frames at 20~fps from each video clip, resulting in the average extracted trajectory length of 33.17 frames. For model training and evaluation, trajectories longer than 20 frames are cropped to 20 frames in length due to the token sequence length limit, resulting in the average length being 15.39 frames.

\myparagraph{Baselines.}
We employ a traditional transformer architecture with a multi-layer perceptron head (Seq2Seq) that recursively predicts the next pose of trajectories~\cite{attention-is-all, rot-cont}. 
This architecture is often utilized for vehicle and pedestrian trajectory forecasting such as~\cite{traj-transformer}. Further details are in the Appendix.
We also apply a 3D hand trajectory forecasting model of USST~\cite{3d-hand-traj}.
USST is an uncertainty-aware state space Transformer model that utilizes the attention mechanism and aleatoric uncertainty. It generates future hand positions in 3D space from a past temporal sequence of the RGB frames and 3D hand positions. USST is different from our VLM-based models in several major points. First, \textit{our models use action descriptions} while USST does not. Instead of action descriptions, USST depends on the past visual features. Second, USST can predict hand positions considering the center of the hand but not rotations, while \textit{ours can predict both the positions and rotations of the objects.} As USST is the hand position prediction model, we use object poses instead of hand poses for our task.

\myparagraph{Evaluation Metrics.}
We evaluate model performance using the following three metrics.
\begin{itemize}
    \item \textbf{Average Displacement Error (ADE)}: ADE is the $l_2$ distance of points in each time step between the predicted and ground-truth trajectories.
    \item \textbf{Final Displacement Error (FDE)}: FDE is the $l_2$ distance of the final points between the predicted and ground-truth trajectories.
    \item \textbf{Geodesic Distance (GD)}: GD~\cite{geodesic-distance} is the angular difference between two rotation matrices as: 
    \begin{equation*}
        d(R^*, R) = \text{arccos}\Big( \frac{tr(R^{*T} R) - 1}{2} \Big), 
    \end{equation*}
    where $R^*$ and $R$ denote the ground-truth and predicted 3D rotations, respectively. $tr$ denotes the trace of matrix.
\end{itemize}
When generating trajectories, some VLM models can halt further generation when they predict a stop token. While this feature allows models to generate arbitrary lengths of trajectories that are not possible for the existing models as of USST~\cite{3d-hand-traj}, this causes a problem in the evaluation due to the varying trajectory length.
Therefore, we apply padding or cutting up for the final pose of a generated trajectory to fit the length of the ground-truth for the evaluation.

\begin{table}[t]
\centering
\vspace*{-2mm}
{
\scriptsize
\begin{tabular}{lccrrrr}
\toprule
 & & & \multicolumn{2}{c}{\textit{3D pos.}} & \multicolumn{2}{c}{\textit{2D pos.}} \\
 \cmidrule(rl){4-5}\cmidrule(rl){6-7}
 & Pre-trained & \datasetname & ADE & FDE & ADE & FDE \\
\midrule
  USST~\cite{3d-hand-traj} & & \checkmark & 0.319 & 0.464 & 0.225 & 0.315 \\
  USST~\cite{3d-hand-traj} & \checkmark & & 0.303 & \textbf{0.414} & 0.243 & 0.310 \\
  USST~\cite{3d-hand-traj} & \checkmark & \checkmark & \textbf{0.254} & 0.448 & \textbf{0.186} & \textbf{0.307} \\
\bottomrule

\end{tabular}
}
\vspace{-2mm}
\caption{
\textbf{Comparison with existing 3DoF trajectory forecasting methods.}
``Pre-trained'' means the model is trained with densely recorded  datasets of H2O~\cite{h2o} and EgoPAT3D~\cite{egopat3d}.
}
\vspace{-4mm}
\label{tab:usst-results}
\end{table}

\begin{figure*}[t]
  \centering
  \includegraphics[width=\textwidth]{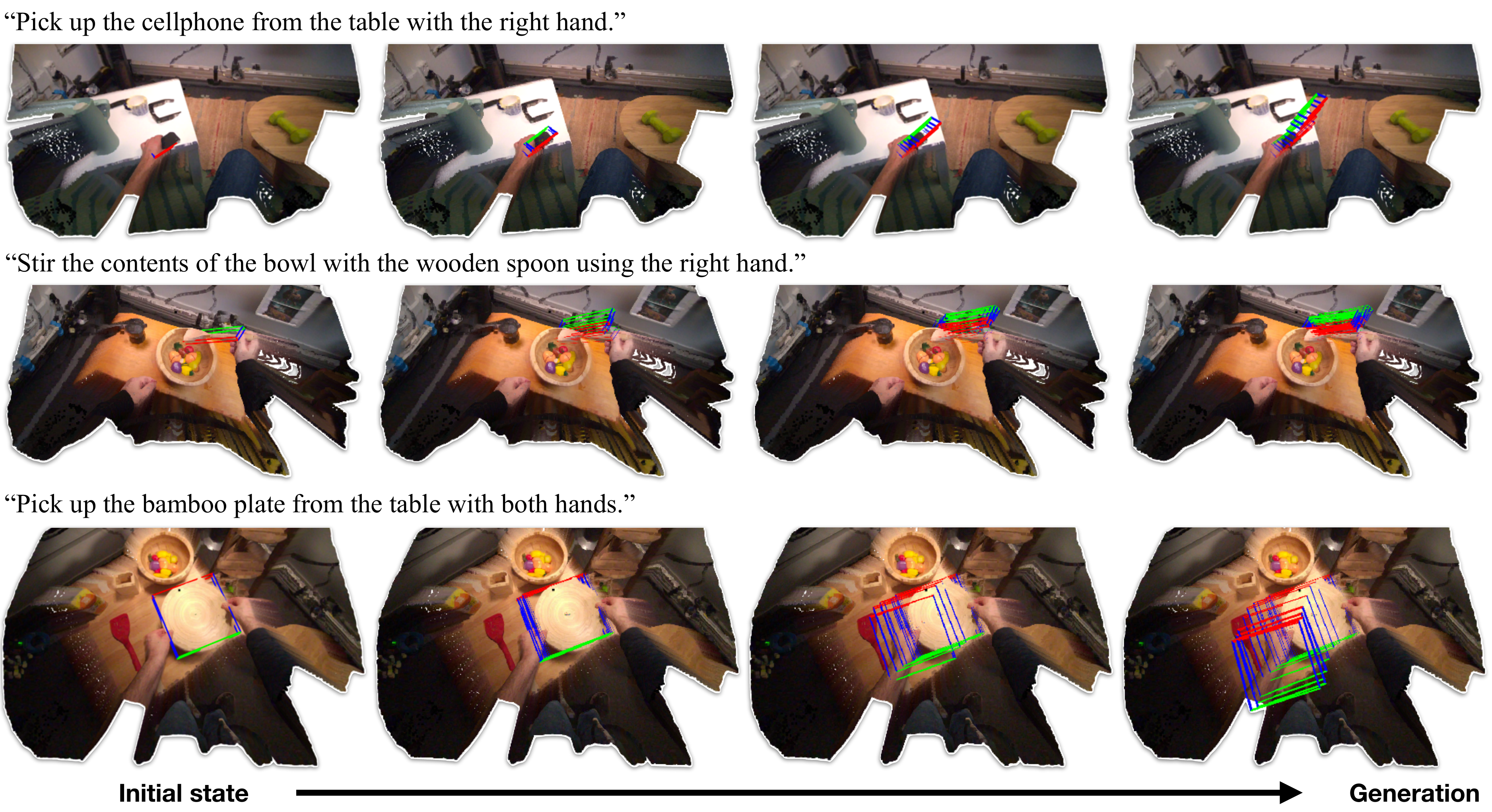}
  \vspace{-4mm}
  \caption{\textbf{Qualitative results of PointLLM~\cite{pointllm}.} Generated trajectories are illustrated using 3D bounding boxes for visualization.}
  \vspace{-5mm}
  \label{fig:qualitative}
\end{figure*}

\subsection{Results}
\cref{tab:results} presents the results of text-guided 3DoF and 6DoF object manipulation trajectory generation in HOT3D. 
VLM-based models outperform conventional Seq2Seq methods. Specifically, when comparing PointLLM to the Seq2Seq (\Bdma{v}) model, PointLLM achieves a reduction of 0.103\,m in ADE (3D). These results demonstrate the effectiveness of utilizing VLMs for object manipulation trajectory generation.

Analysis of VLM-based models reveals two key insights. First, when comparing model performance across different visual modalities, the utilization of point clouds significantly contributes to performance improvement. Additionally, the use of depth maps in 2D-VLMs tends to result in better performance. These results suggest that incorporating spatial information such as depth and point clouds boosts model performance.
Second, although general scaling laws suggest that larger LLMs achieve better performance, this trend does not consistently apply to trajectory prediction. For instance, BLIP-2 (2.7B) outperforms VILA (8B). 
These results suggest that improving model performance for our task requires not only increasing model size but also considering the specific characteristics of each LLM.

\myparagraph{Comparison with Existing Trajectory Set.}
\cref{tab:usst-results} presents the performance of the USST models~\cite{3d-hand-traj}.
The USST model that is trained with our extracted trajectories from scratch performed on-par with the USST model that is pre-trained with densely recorded datasets~\cite{egopat3d, h2o}. When we further fine-tuned the pre-trained USST model with our dataset, we observed the performance gain in many metrics, confirming the effectiveness of the automatically extracted trajectories.

\begin{table}[t]
    \centering
    {
    \footnotesize
    \begin{tabular}{crrrrr}
    \toprule
    & \multicolumn{2}{c}{\textit{3D pos.}} & \multicolumn{2}{c}{\textit{2D pos.}} & \multicolumn{1}{c}{\textit{3D rot.}}\\
     \cmidrule(rl){2-3}\cmidrule(rl){4-5}\cmidrule(rl){6-6}
    Sampling & \multicolumn{1}{c}{ADE} & \multicolumn{1}{c}{FDE} & \multicolumn{1}{c}{ADE} & \multicolumn{1}{c}{FDE} & \multicolumn{1}{c}{GD} \\
    \midrule
    1  & 0.271 & 0.458 & 0.208 & 0.327 & 0.541  \\
    3  & 0.236 & 0.401 & 0.178 & 0.280 & 0.543 \\
    10 & \textbf{0.212} & \textbf{0.363} & \textbf{0.158} & \textbf{0.252} & \textbf{0.540} \\
    \bottomrule
    \end{tabular}
    }
    \vspace{-2mm}
    \caption{
    \textbf{Performance comparison in probabilistic sampling for PointLLM~\cite{pointllm}.} ``Sampling'' column indicates the number of samples generated. 
    }
    \vspace{-3mm}
    \label{tab:sampling}
\end{table}

\myparagraph{Probabilistic Sampling.}
There are possible multiple trajectories that match the given action description~\cite{Choi_2019_ICCV, Lee_2017_CVPR}.
It is difficult for existing models to generate multiple possible trajectories as USST does not rely on action descriptions.
Fortunately, our model uses the language modeling approach that allows generating multiple different trajectories using nucleus sampling~\cite{nucleus-sampling}.
We employ PointLLM as it demonstrated the best performance among the baselines for this purpose. For each instance, we report the lowest ADE (3D) score from the sampled trajectories.
\cref{tab:sampling} shows that increasing the number of samples improves model performance. These results imply that our model successfully generates multiple trajectory candidates to perform the given action descriptions. Meanwhile, the performance improvement for GD is smaller than other positional metrics, indicating the difficulty in generating accurate rotational movements.

\begin{figure}[t]
  \centering 
  \vspace{-2mm}
  \includegraphics[width=0.95\linewidth]{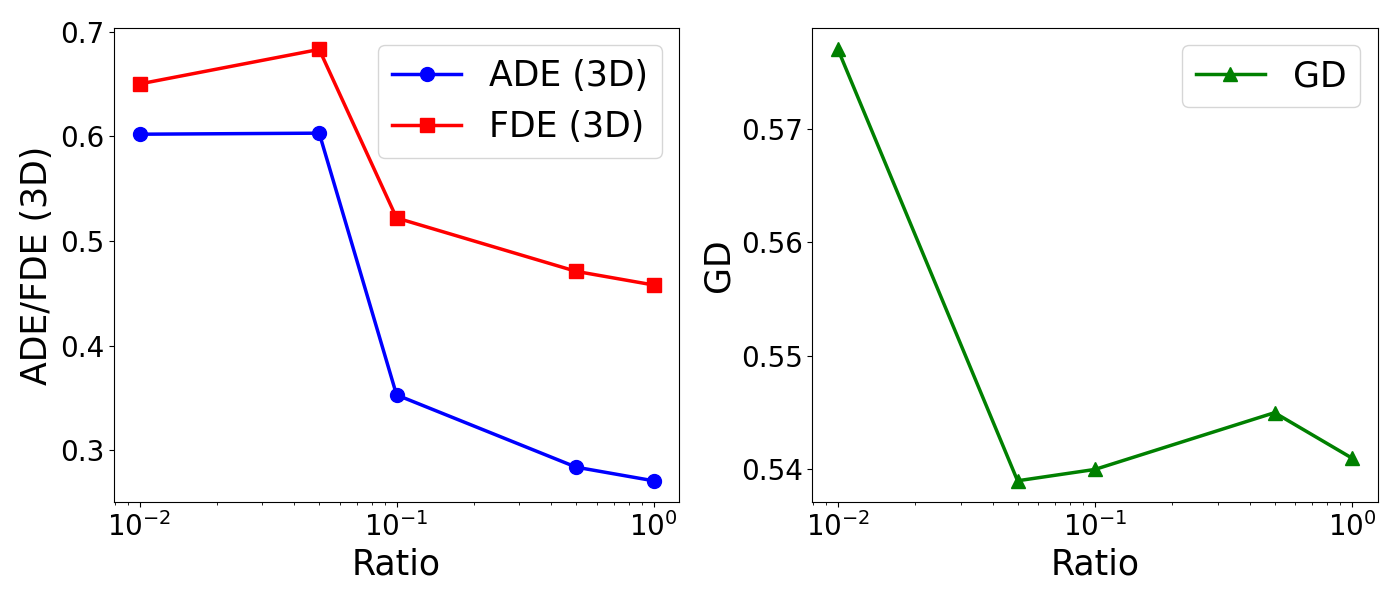}
  \vspace{-4mm}
  \caption{\textbf{Comparison of performance across different dataset scales for PointLLM~\cite{pointllm}.}}
  \vspace{-3mm}
  \label{fig:scale-graph}
\end{figure}
\myparagraph{Data Scalability.}
As our framework automatically generates manipulation trajectories, we investigate the impact of dataset scale on model performance.
We utilized training data at ratios of 1.0, 0.5, 0.1, 0.05, and 0.01 of the full dataset to assess performance across different dataset scales.
\cref{fig:scale-graph} presents the performance comparison under varying dataset scales.
We use PointLLM for this purpose as it achieved the highest performance in \cref{tab:results}.
The results, especially in ADE and FDE, indicate that model performance improves as the dataset size increases, demonstrating the effectiveness of our dataset creation framework. 

\myparagraph{Qualitative Results.}
\begin{figure}[t]
  \centering
  \includegraphics[width=\linewidth]{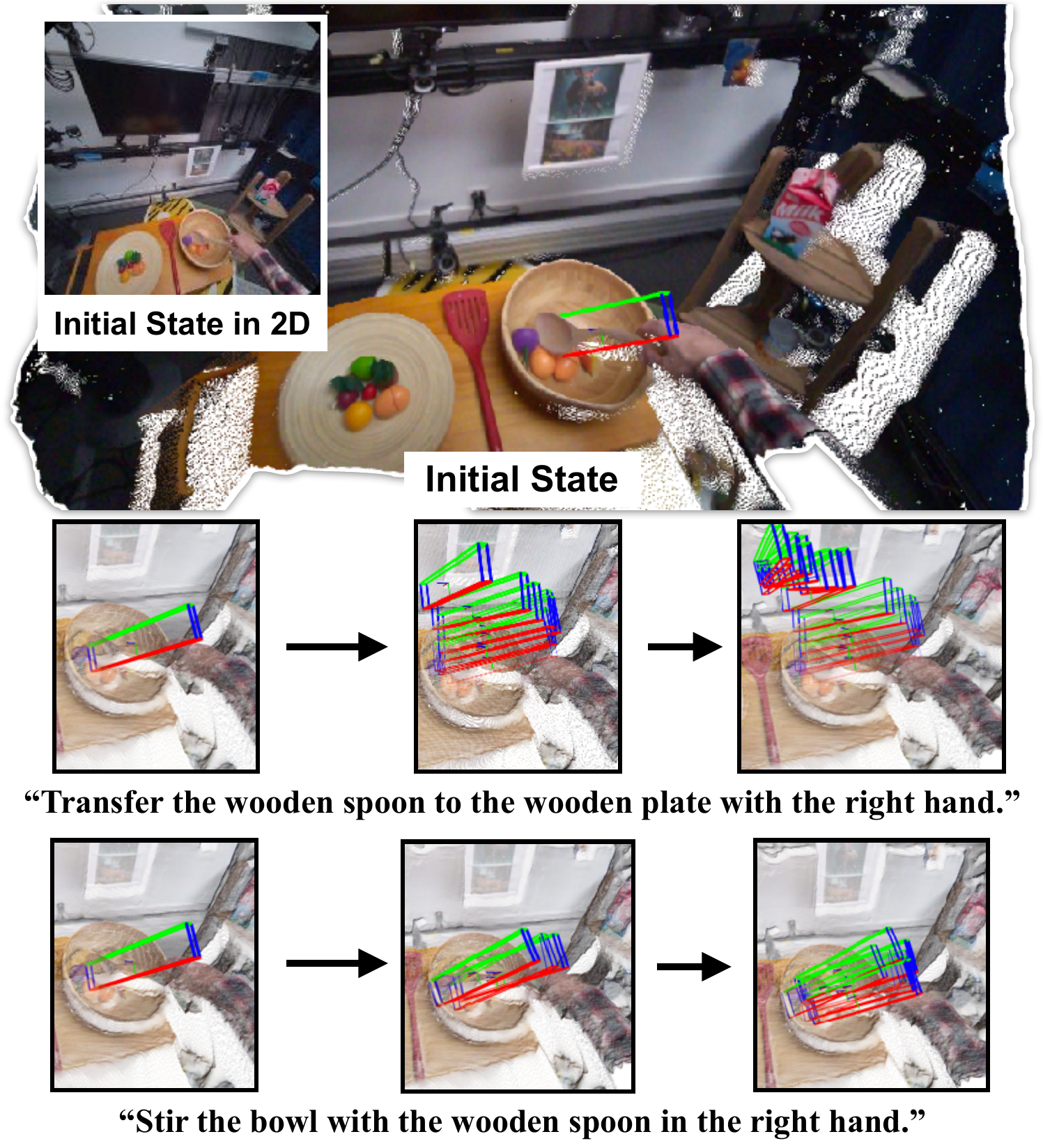}
  \vspace{-3mm}
  \caption{\textbf{PointLLM~\cite{pointllm} results with different action descriptions.} 
  The generations are visualized with 3D bounding boxes.} 
  \vspace{-1mm}
  \label{fig:qualitative-demo}
\end{figure}
\cref{fig:qualitative} illustrates the generated trajectories of PointLLM along with the changes of the 6DoF object pose from the initial pose.
It is interesting that regardless of the difference of the objects in HOT3D from those in the Ego-Exo4D videos, our method successfully generates understandable object manipulation trajectories depending on action descriptions and their verbs.
For example, the case in the middle row is with the verb ``stir'' and it yields a different trajectory from other examples.

\myparagraph{Action Description Change.}
We survey the impact of the action descriptions on the generation of trajectories, as it is highly expected that the generated trajectories are modified depending on the action descriptions. 
\cref{fig:qualitative-demo} presents two generated trajectories from the trained PointLLM, confirming two different movements depending on the action descriptions: the wooden spoon is pulled away from the plate in the top row while it stays on the bowl and is being shaken on the bottom row.
Further visualizations including demonstration videos in our project page.

\begin{table}[t]
\centering
\footnotesize
\vspace{-1mm}
\begin{tabular}{lcrrr}
\toprule
 & Trajectory & \multicolumn{1}{c}{$\text{Sim}_v$} & \multicolumn{1}{c}{METEOR} & \multicolumn{1}{c}{BLEU-4}\\
\midrule
\multirow{2}{*}{BLIP-2~\cite{blip2}}
 & &  58.59 & 23.76 & 11.93 \\
 & \checkmark & \textbf{67.58} & \textbf{24.19} & \textbf{12.48} \\
\midrule
\multirow{2}{*}{PointLLM~\cite{pointllm}}
 & &  43.05 & 17.42 & 3.25 \\
 & \checkmark & \textbf{53.33} & \textbf{17.56} & \textbf{4.44}\\
\bottomrule
\end{tabular}
\vspace{-2mm}
\caption{
\textbf{Effect of trajectory information on image and 3D captioning performance} The ``Trajectory'' column indicates whether trajectory information is utilized by the model (\checkmark). Higher values indicate better performance across all metrics.
}
\vspace{-3mm}
\label{tab:caption-results}
\end{table}
\begin{figure}[t]
  \centering
  \includegraphics[width=\linewidth]{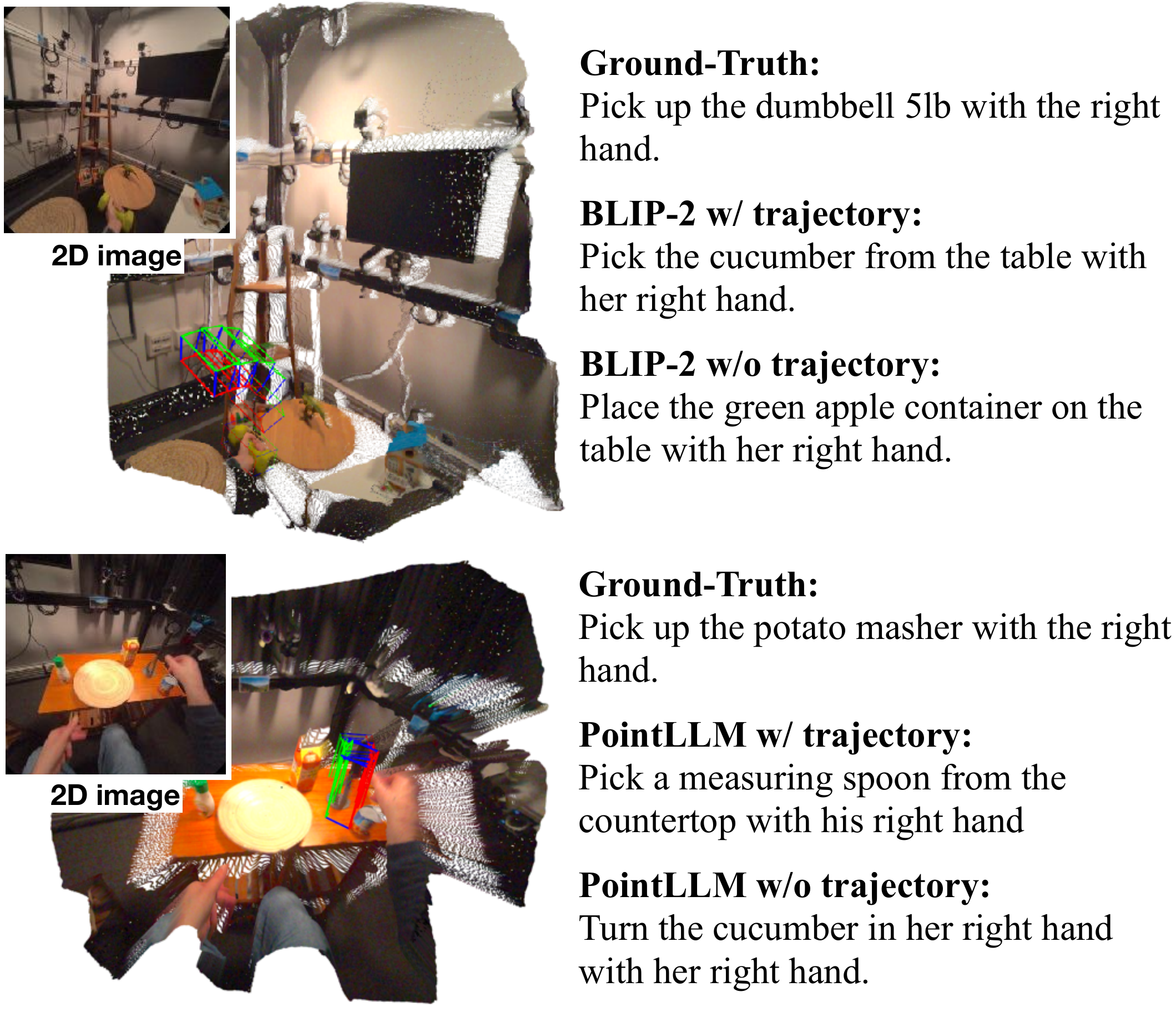} 
  \vspace{-8mm}
  \caption{\textbf{Qualitative results in image and 3D captioning task.} The top and bottom figures depict generated captions of BLIP-2~\cite{blip2} and PointLLM~\cite{pointllm}, respectively. ``w/'' and ``w/o'' indicate whether models utilize trajectory information.}
  \vspace{-3mm}
  \label{fig:qualitative-caption}
\end{figure}

\subsection{Action Description Generation}
Finally, we conduct experiments of generating action descriptions with BLIP-2 and PointLLM. Indeed, current VLMs are so effective that they can learn to generate at first-look plausible action descriptions from solely 2D image 3D scenery inputs. Therefore, we test two conditions of with and without trajectories.
In addition to the standard metrics of METEOR~\cite{meteor} and BLEU-4~\cite{bleu} for action description evaluation, we also report the similarity of the verbs between the generated and annotated trajectories, $\text{Sim}_{v}$, because they reflect the differences in the actions by camera wearers in egovision. $\text{Sim}_v$ is computed as the cosine similarity of word embeddings~\cite{glove}.
\cref{tab:caption-results} shows the results from BLIP-2~\cite{blip2} and PointLLM~\cite{pointllm}. The results indicate that utilizing trajectories boosts performance, particularly verb similarity, by nearly 10\% in both image and 3D captioning, confirming the effectiveness of our framework for extracting accurate manipulation trajectories that reflect human motion. 
\cref{fig:qualitative-caption} illustrates the qualitative results of image and 3D captioning, indicating that object motion trajectories enhance verb similarity in both models.
\section{Conclusion}
We developed a scalable framework to extract 6DoF object manipulation trajectories from egocentric videos. 
Using these trajectories, we created generation models for object manipulation trajectories based on visual and point cloud-based language models. 
Our experimental results on HOT3D dataset demonstrate that our models learn to generate object trajectories from action descriptions, providing fair baseline models for the new task. 
Moreover, utilizing trajectories for visual captioning tasks reveals that our dataset creation framework successfully extracts meaningful trajectories from egocentric videos. For future work, we construct a massive-scale dataset by adapting the proposed approach to other egocentric video datasets and investigate its effectiveness in robotic manipulation.

\myparagraph{Limitations.}
The current framework has two limitations. First, because our framework defines an object pose as a 6DoF bounding box, it cannot be applied to deformable objects such as clothes. Second, we observed failure cases in our dataset due to object segmentation and point-cloud registration. Segmentation can fail when similar objects are present, while registration can fail when camera pose changes abruptly. 
Further details are in the Appendix.

\section*{Acknowledgments}
This work was supported by JSPS KAKENHI Grant Number JP22K17983, JP22KK0184 and JST CRONOS JPMJCS24K6.

{
    \small
    \bibliographystyle{ieeenat_fullname}
    \bibliography{main}
}

\appendix
\clearpage
\setcounter{page}{1}
\maketitlesupplementary

We provide detailed descriptions of our framework, dataset statistics, a baseline of Seq2Seq implementation, and implementation details of vision and point cloud-based language models. We also provide visualization of extracted trajectories and qualitative results of models in the supplementary video.

\section{Dataset}
\begin{figure*}[t]
  \centering
  \includegraphics[width=0.9\textwidth]{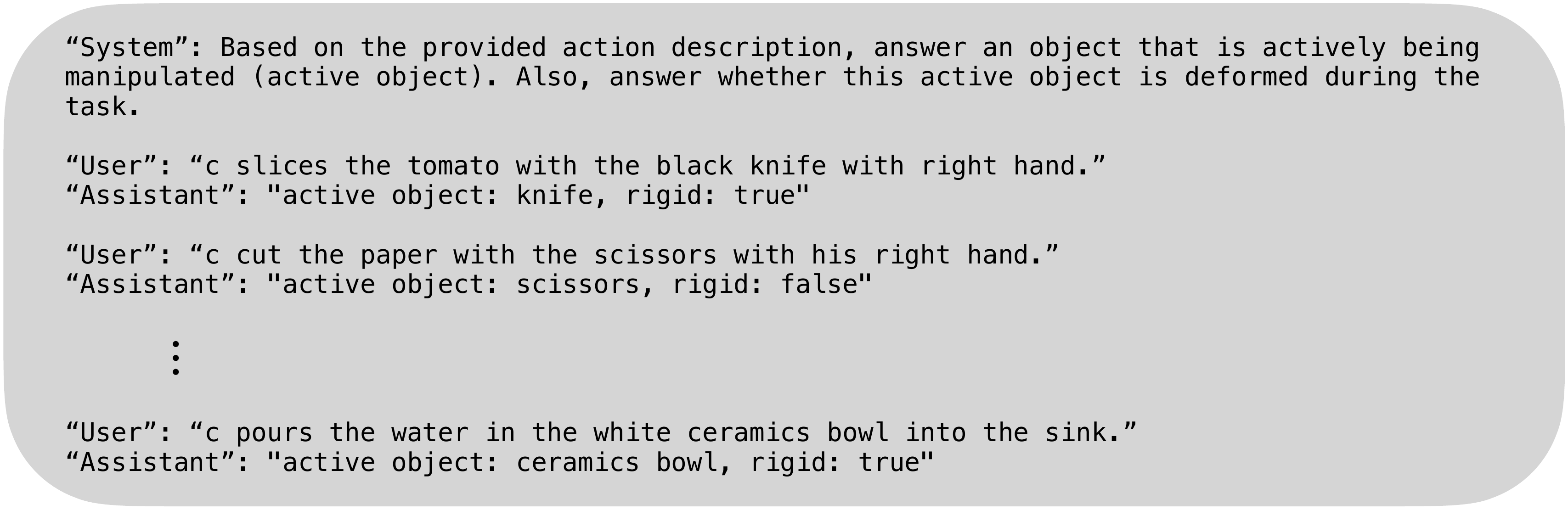}
  \caption{\textbf{System prompt to obtain manipulated objects.}}
  \label{fig:prompt-for-mon}
\end{figure*}
\begin{figure*}[t]
  \centering
  \includegraphics[width=0.9\textwidth]{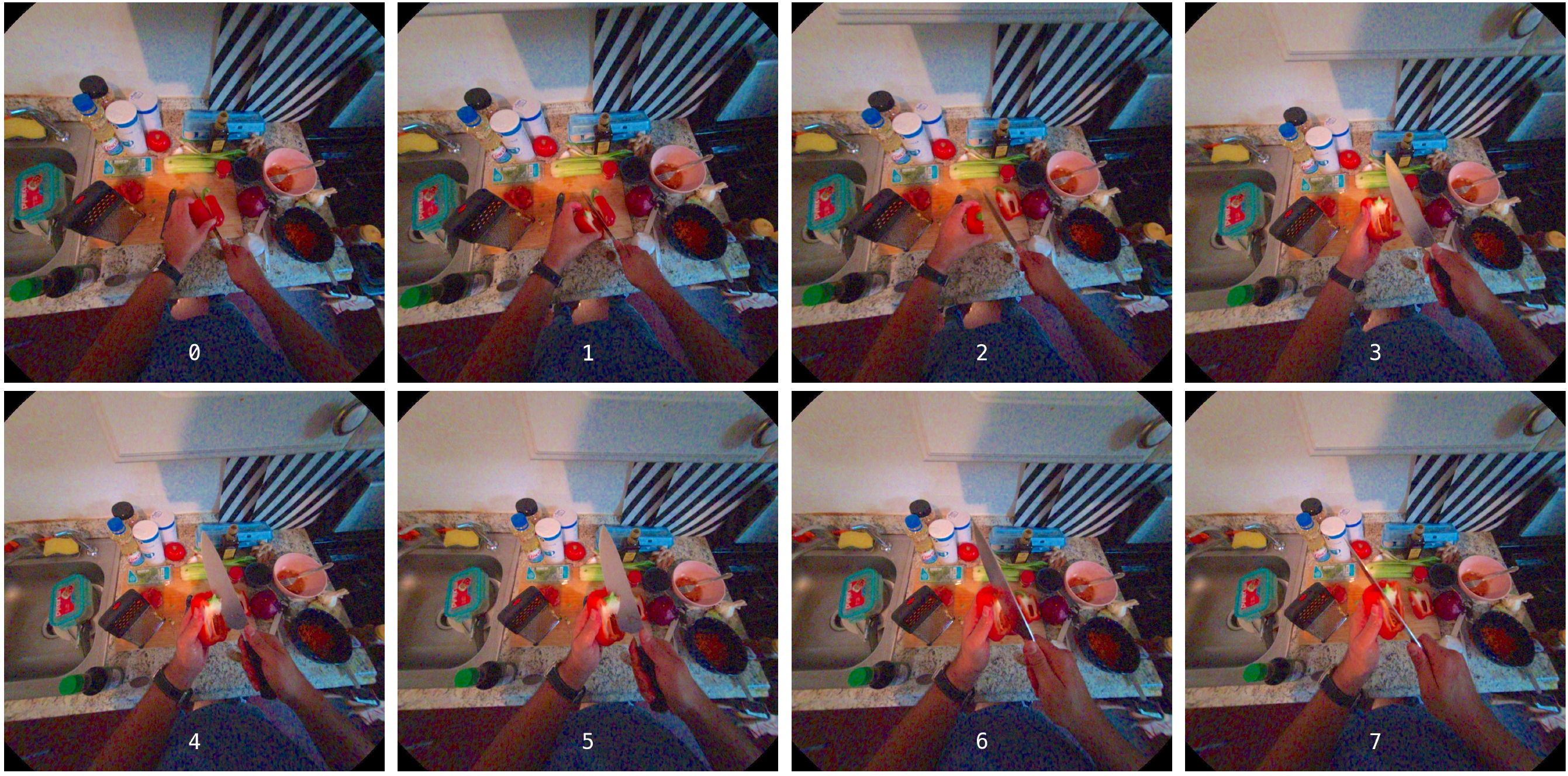}
  \caption{\textbf{Image frames embedded sequential indices for GPT-4o input.}}
  \label{fig:embed-imgs}
\end{figure*}
\begin{figure*}[t]
  \centering
  \includegraphics[width=0.9\textwidth]{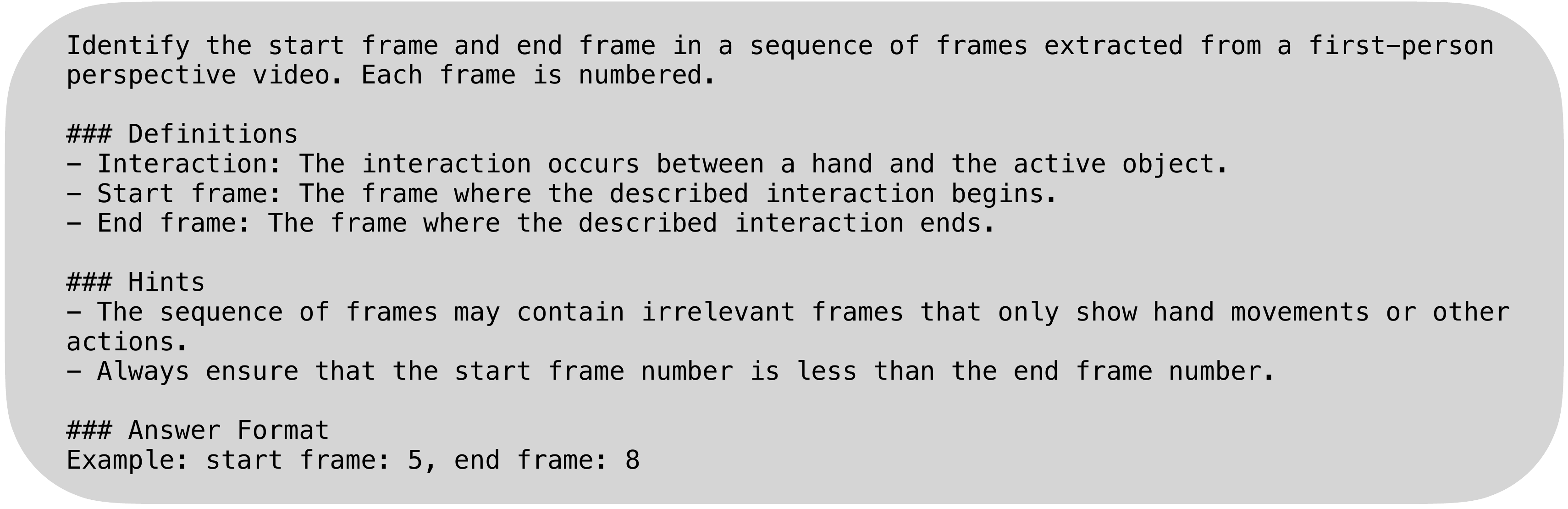}
  \caption{\textbf{System prompt for temporal action localization.}}
  \label{fig:prompt-for-tal}
\end{figure*}
\subsection{Training Data}
In addition to the descriptions in the main paper, we describe the details of temporal action localization, position sequence traction, and trajectory projection.

\myparagraph{System Prompt for Temporal Action Localization.}
To obtain a manipulated object name, determine whether it is rigid, and localize the start and end timestamps of an action from a video clip, we use OpenAI GPT-4o~\cite{gpt-4} in two stages. First, to obtain the name of the manipulated object and determine whether the manipulated object is rigid or not, we conduct few-shot learning. We provide an action description and a system prompt containing several samples to GPT-4o without visual input. Figure~\ref{fig:prompt-for-mon} shows the system prompt and several samples for few-shot learning for this task. Second, we provide image frames assigned sequential indices, an action description, the manipulated object name, and a system prompt to GPT-4 to determine the start and end timestamps of the action for each video clip. The maximum sequence of image frames is eight, and embedded frames are depicted in Figure~\ref{fig:embed-imgs}. Figure~\ref{fig:prompt-for-tal} shows the system prompt for this task.

\myparagraph{Details of Position Sequence Extraction.}
To obtain segmentation maps of manipulated objects, we utilize the open-vocabulary segmentation model~\cite{g-dino} as discussed in the main paper. However, egocentric videos of daily activities often involve the same objects within scenes, leading to incorrect object segmentation maps. To mitigate this issue, we employ a hand-object detection model~\cite{hod} to detect interacted objects across all frames. We calculate the intersection-over-union (IoU) between the detected object bounding boxes and object segmentation map candidates, selecting the segmentation map with the highest IoU as the manipulated object segmentation map. Additionally, we filter out a video clip if the confidence score of the detection results falls below a threshold of 0.3.

\myparagraph{Details of Trajectory Projection.}
To obtain projection matrices between image frames, we perform RANSAC-based global registration and colored iterative closest point (ICP) algorithm. For RANSAC-based global registration, we set the distance threshold to 0.03, the maximum number of iterations to 100{,}000, and the confidence level to 0.999. For the colored ICP algorithm, we set the distance threshold to 0.008, the maximum number of iterations to 100, the relative fitness to $1 \times 10^{-6}$, and the relative root mean square error to $1 \times 10^{-6}$.

\begin{table}[t]
    \centering
    {
    \footnotesize
    \begin{tabular}{lrrrr}
    \toprule
    & \multicolumn{1}{c}{Music} & \multicolumn{1}{c}{Cooking} & \multicolumn{1}{c}{Bike Repair} & \multicolumn{1}{c}{Health Care} \\
    \midrule
    \# Trajectories & 816 & 20,529 & 1,448 & 5,704 \\
    \# Frames & 47,045 & 640,487 & 53,270 & 202,731 \\
    \midrule
    avg. Frames & 58 & 31 & 38 & 36 \\
    \bottomrule
    \end{tabular}
    }
    \vspace{-2mm}
    \caption{
    \textbf{Statistics of extracted trajectories in each scenario.}
    ``avg.'' stands for average.
    }
    \vspace{-2mm}
    \label{tab:stats-scenario}
\end{table}

\myparagraph{Details of Resource Dataset.}
To construct our dataset, we utilize the Ego-Exo4D~\cite{egoexo4d} dataset, which encompasses eight diverse scenarios: dance, soccer, basketball, bouldering, music, cooking, bike repair, and health care. To focus on object interaction and ensure a stable trajectory projection process, we filter out the scenarios involving dance, soccer, basketball, and bouldering. \cref{tab:stats-scenario} presents the number of trajectories and frames of trajectory for each scenario. The average number of frames of trajectory is calculated by dividing the total number of frames by the number of trajectories.
As shown in \cref{tab:stats-scenario}, the cooking scenario constitutes the majority of our dataset. This result may be attributed to two reasons. First, the Ego-Exo4D dataset originally includes more instances of the cooking scenario compared to other scenarios. Second, objects in the bike repair and health care scenarios, such as a COVID-19 test plate, are more challenging to detect than those in the cooking scenario. Consequently, video clips from these scenarios are automatically filtered out.
Moreover, the average number of frames per trajectory in the music scenario is higher than in other scenarios. This may be due to the characteristics of musical activities, playing musical instruments typically longer than other actions such as ``grab a cup.’'

\begin{figure}
    \centering
    \includegraphics[width=\linewidth]{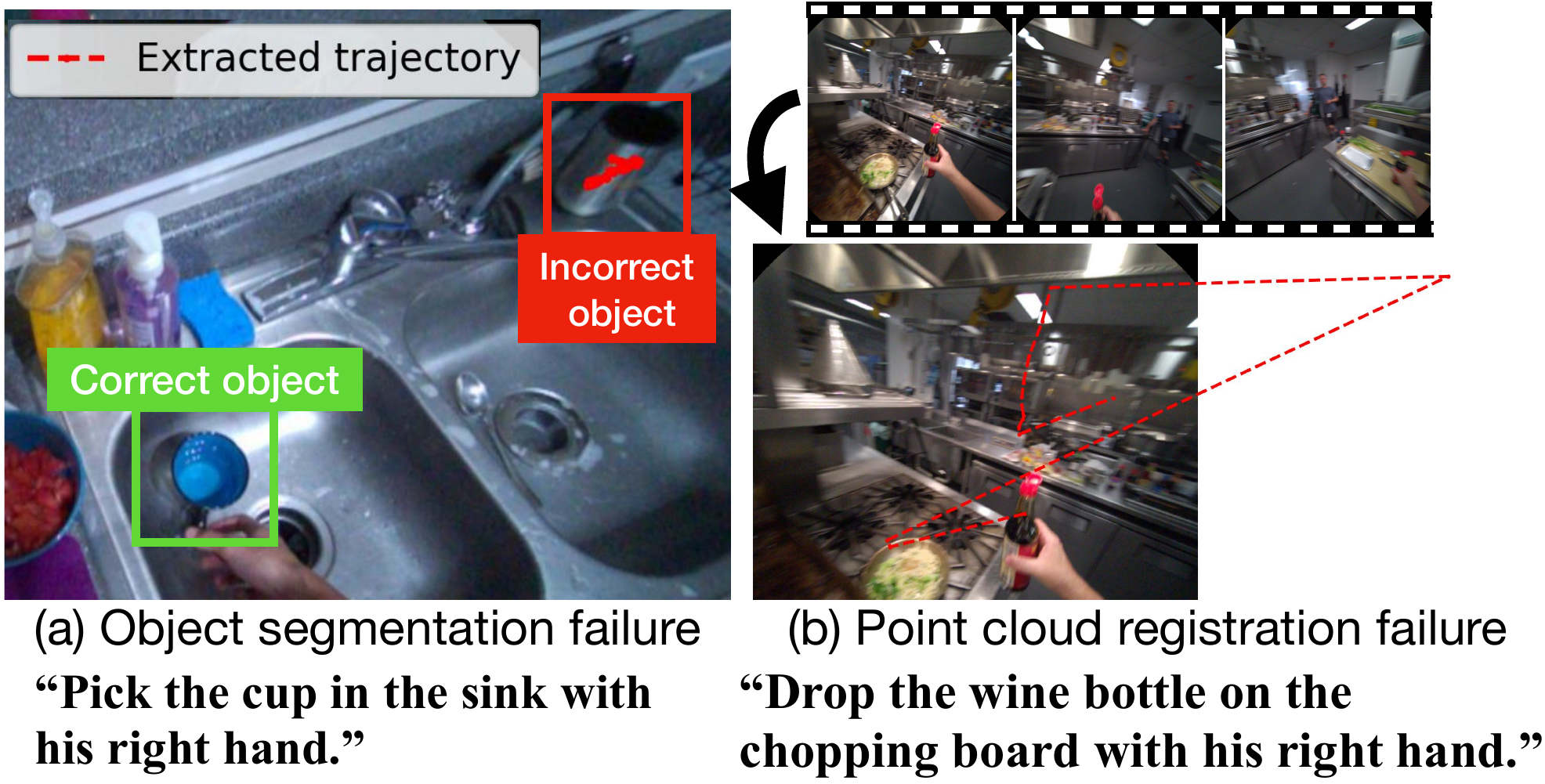}
    \caption{\textbf{Failure cases.}}
    \label{fig:failure-case}
\end{figure}

\myparagraph{Failure Cases.}
There are unavoidable failure cases and we have carefully filtered out such cases through data curation methods. \cref{fig:failure-case} illustrates failures from object segmentation and point cloud registration during the annotation process.
Object segmentation can fail when multiple similar objects are present, while registration can fail when camera pose changes abruptly. 

\myparagraph{Data Curation Methods.}
In our study, we applied two filtering steps to remove inaccurate trajectories. 
First, we excluded incorrect segmentation results using a hand-object detector~\cite{hod}, such as \cref{fig:failure-case} (a). 
Second, we removed trajectories that were out of frame in observation images, such as \cref{fig:failure-case} (b). 

\begin{figure*}[t]
  \centering
  \includegraphics[width=0.9\textwidth]{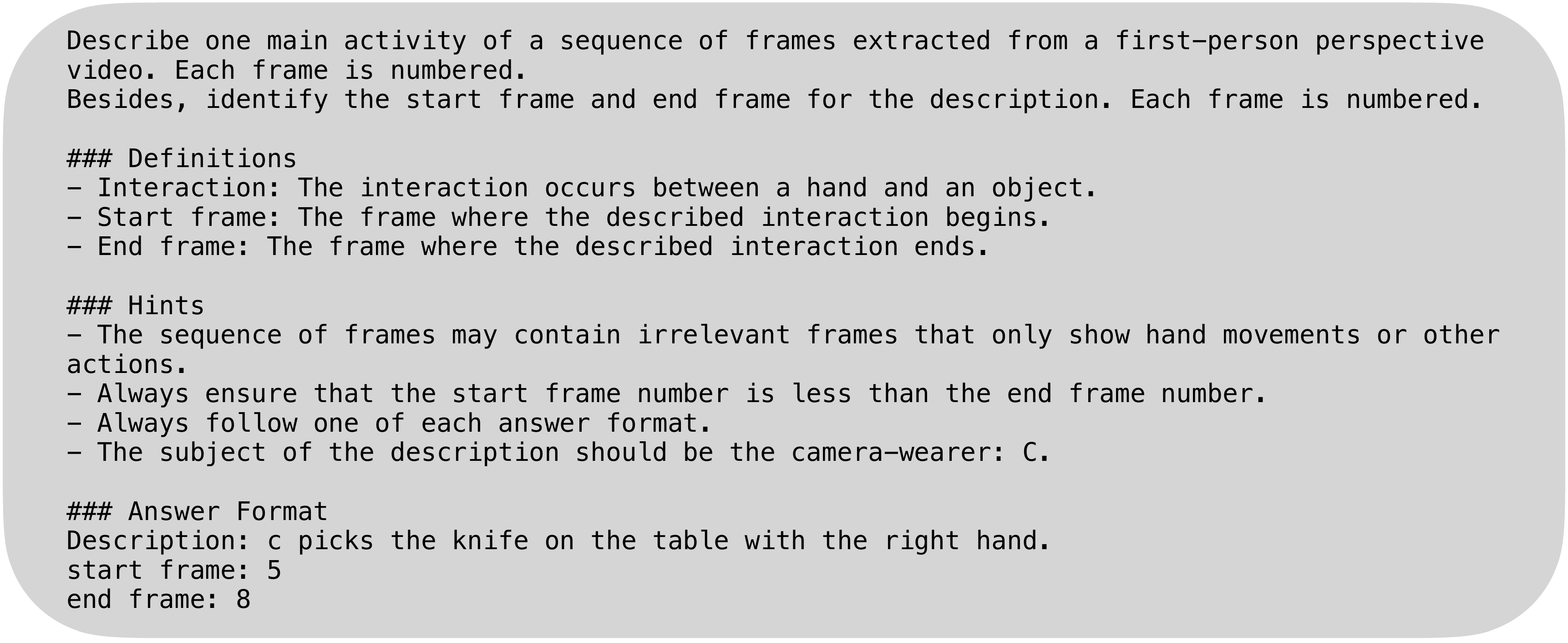}
  \caption{\textbf{System prompt for HOT3D annotation.}}
  \label{fig:prompt-for-hot3d}
\end{figure*}
\subsection{Evaluation Data}
In addition to the descriptions in the main paper, we describe how to determine manipulated objects within scenes, and how to annotate action descriptions using OpenAI GPT-4o~\cite{gpt-4}.

\myparagraph{Manipulated Objects Determination.}
To extract object manipulation trajectories, we need to detect which objects within a scene are being manipulated. To achieve this, unlike the approach used in constructing the training dataset, we utilize the annotated trajectories of each object provided in the HOT3D~\cite{hot3d} dataset. For each video clip, we compute the displacement of each object and identify the object with the highest displacement as the manipulated object.

\myparagraph{Action Description Generation.}
Since no textual information or action start and end timestamps exist in the HOT3D~\cite{hot3d} dataset, we need to annotate them to align with our task setting. To achieve this, we adopt a similar workflow to the training dataset construction. We first split raw egocentric videos into several video clips, each spanning a four-second interval. Next, we perform temporal action localization and require the model to generate action descriptions. Additionally, we include an object name originally annotated in HOT3D as a user query to guide OpenAI GPT-4o in focusing on actions involving interaction with the object for each instance. \cref{fig:prompt-for-hot3d} depicts the system prompt for this task.

\begin{figure}[t]
  \centering
  \includegraphics[width=\linewidth]{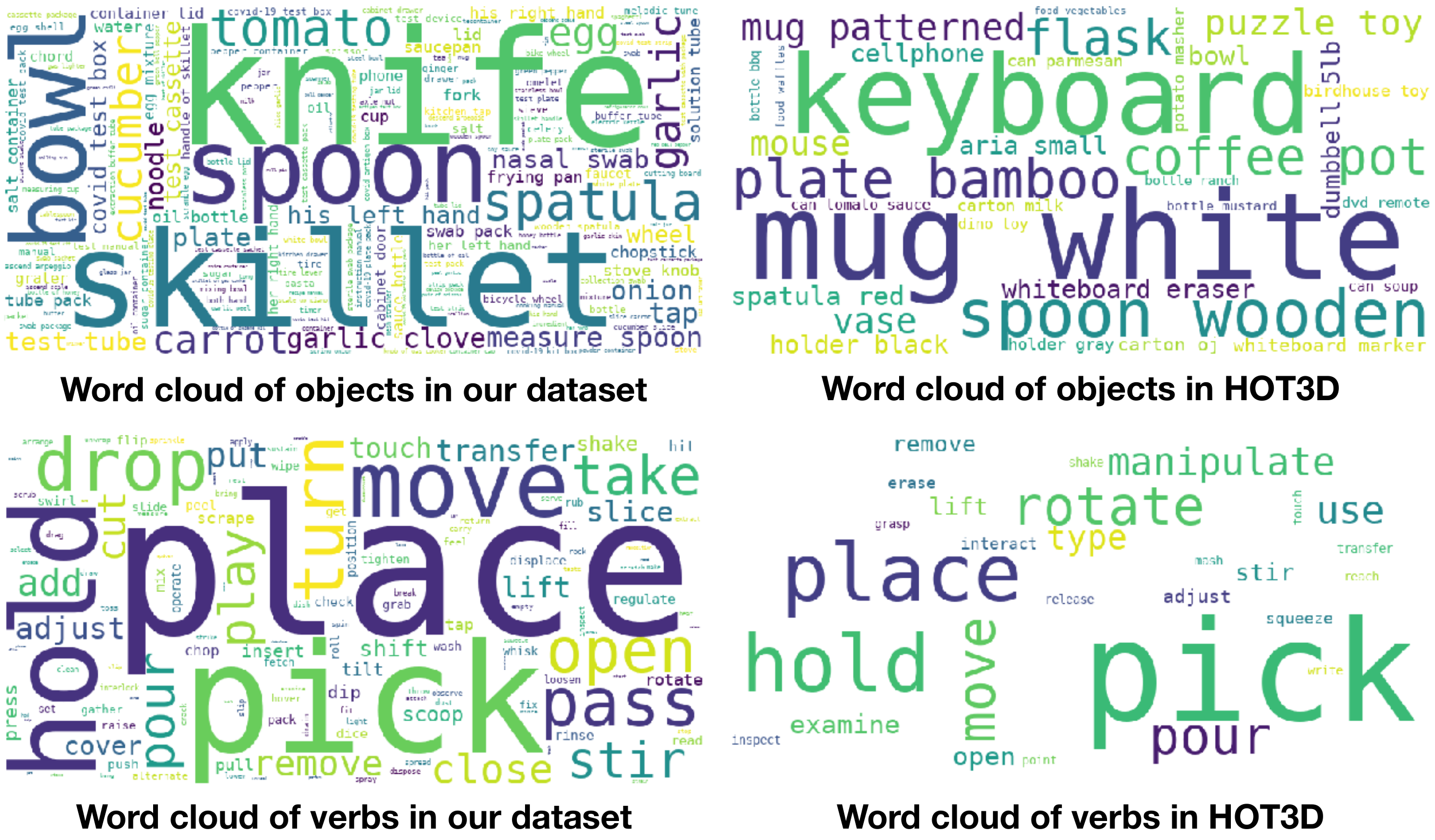}
  \caption{\textbf{Word cloud of objects and verbs in our dataset and HOT3D dataset~\cite{hot3d}.}}
  \label{fig:wordcloud}
\end{figure}
\subsection{Dataset Statistics}
Here, we provide detailed statistics for our dataset and the HOT3D evaluation dataset.

\myparagraph{Vocabularies.}
\cref{fig:wordcloud} depicts word clouds of objects and verbs that appeared in action descriptions for both our dataset and HOT3D dataset. 
Our dataset consists of diverse objects for a wide range of verbs, enhancing models' generalization capability for read-world scenarios. 
Additionally, the objects in our dataset significantly differ from those in HOT3D, indicating that our approach successfully generates manipulation trajectories even for rare or unseen objects.

\begin{figure}[t]
  \centering
  
  \begin{subfigure}[t]{\linewidth}
    \centering
    \includegraphics[width=\linewidth]{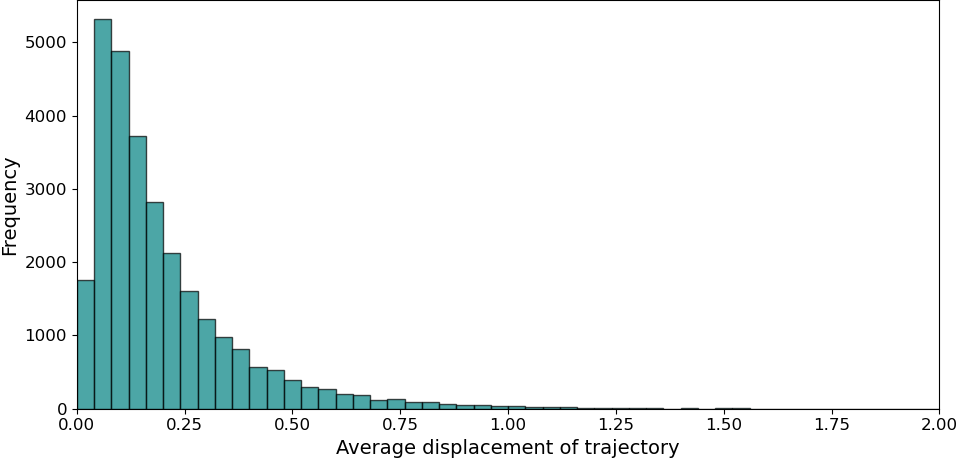}
    \caption{Our dataset}
    \label{fig:displacement-training}
  \end{subfigure}
  
  \vspace{1mm} 
  
  \begin{subfigure}[t]{\linewidth}
    \centering
    \includegraphics[width=\textwidth]{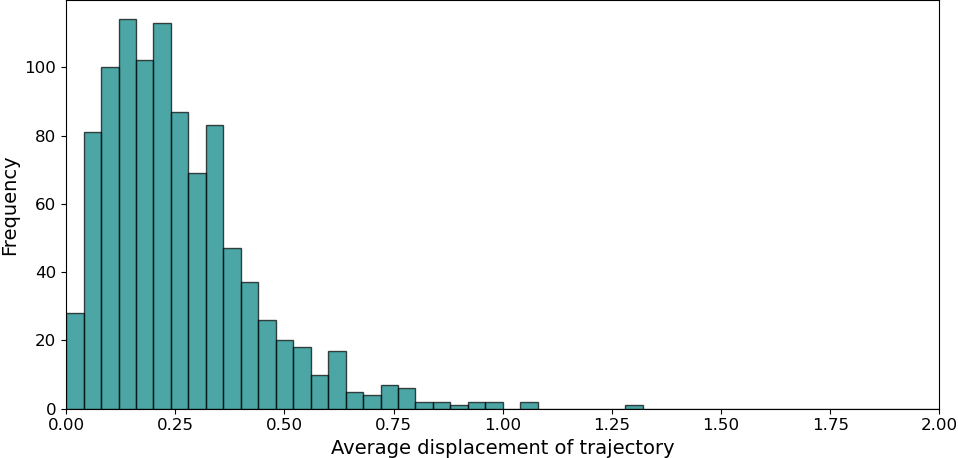}
    \caption{HOT3D dataset}
    \label{fig:displacement-hot3d}
  \end{subfigure}
  
  \caption{\textbf{Distribution of the average displacement of trajectories in our dataset and HOT3D dataset.}}
  \label{fig:displacement}
\end{figure}

\myparagraph{Average Displacement of Trajectories.}
\cref{fig:displacement} illustrates the statistics of average displacement of extracted object manipulation trajectories for both our dataset and HOT3D dataset.
Although our dataset is constructed automatically, it has a similar distribution to that of HOT3D, thereby demonstrating the validity of our approach.

\begin{figure*}[t]
  \centering
  
  \begin{subfigure}[t]{0.9\textwidth}
    \centering
    \includegraphics[width=\textwidth]{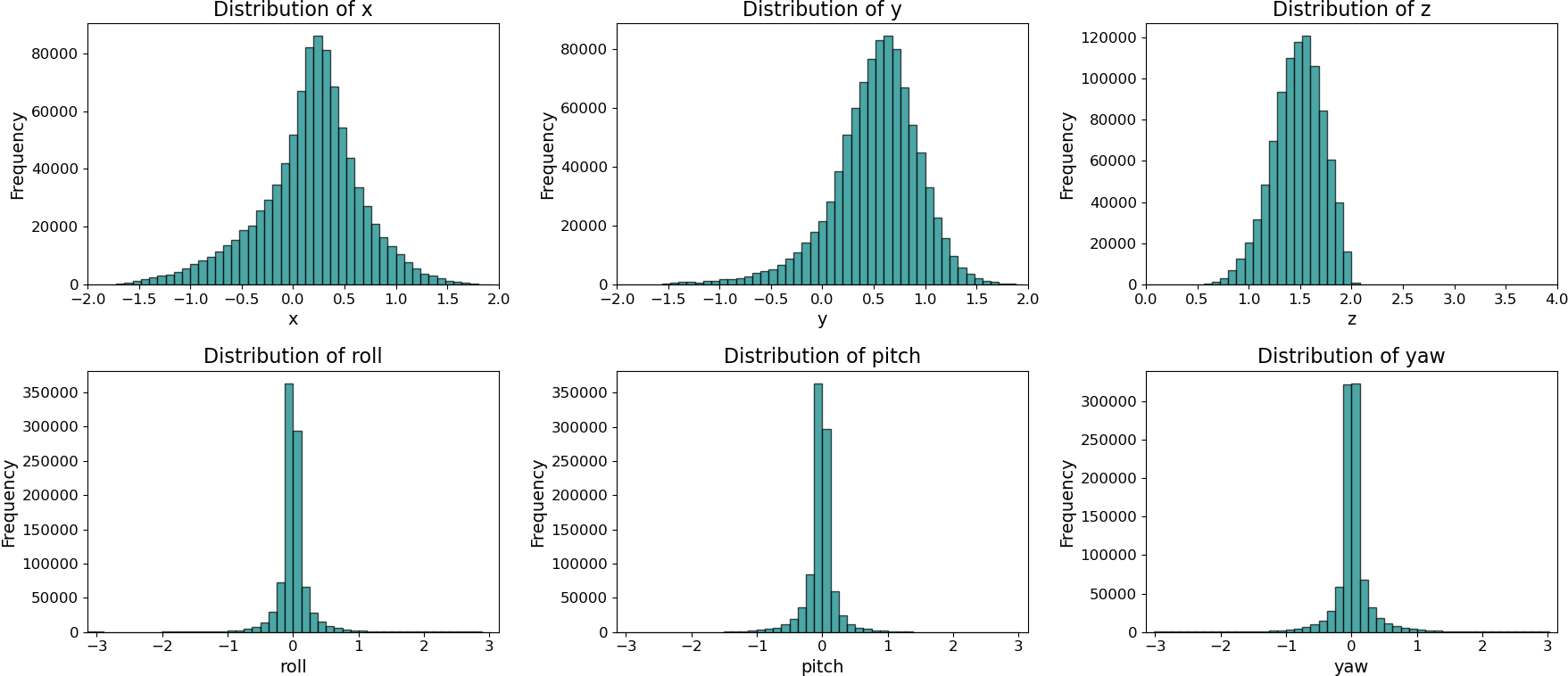}
    \caption{Our dataset}
  \end{subfigure}
  
  \vspace{1mm} 
  
  \begin{subfigure}[t]{0.9\textwidth}
    \centering
    \includegraphics[width=\textwidth]{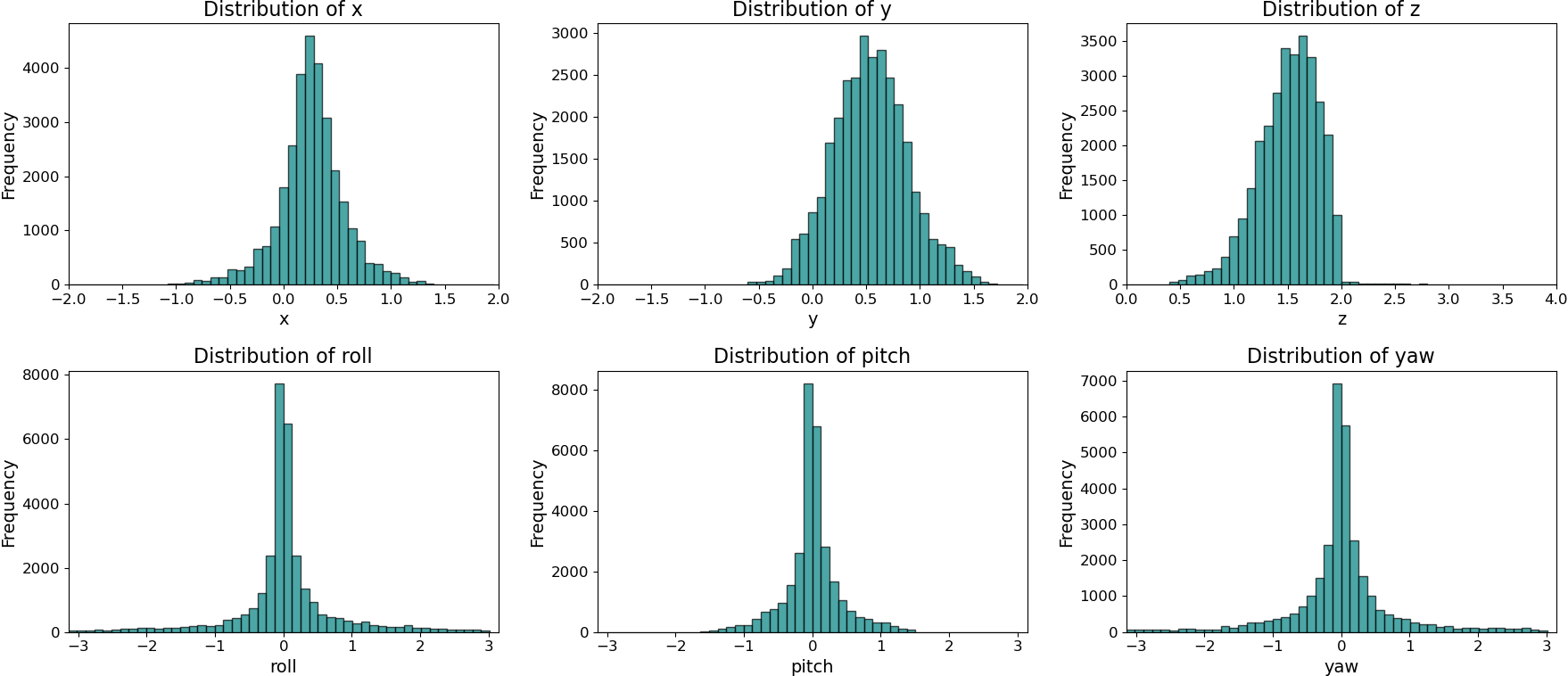}
    \caption{HOT3D dataset}
  \end{subfigure}
  
  \caption{\textbf{Distribution of variations in each trajectory element for our dataset and HOT3D dataset.}}
  \label{fig:variation}
\end{figure*}

\myparagraph{Variation in Each Element.}
\cref{fig:variation} illustrates the distribution of each trajectory element for both our dataset and the HOT3D~\cite{hot3d} dataset. The distribution of each element in our training dataset is similar to that of HOT3D. However, some variations in the rotational elements of our dataset slightly differ from those in HOT3D. These differences may arise from the domain disparity between Ego-Exo4D, the source of our dataset, and HOT3D. While the Ego-Exo4D dataset captures daily activities, HOT3D is designed for object tracking challenges. Consequently, the object manipulation motions in HOT3D involve actions with significant rotational movement, such as object inspection scenarios.

\section{Baseline Seq2Seq Model}
In our experiments, we utilize a Seq2Seq transformer with an MLP head as the baseline model~\cite{attention-is-all}. The transformer comprises four layers, four attention heads, and a hidden size of 256. The MLP outputs a pose represented by $[x, y, z, \text{roll}, \text{pitch}, \text{yaw}]$. To address the issue of rotational continuity~\cite{rot-cont}, we represent each angle $\theta \in \{\text{roll}, \text{pitch}, \text{yaw}\}$ using $[\cos(\theta), \sin(\theta)]$. After this transformation, each element of the parameters is normalized to the range $[0, 1]$.

\section{Implementation Details of Our Model}
We use AdamW~\cite{adamw} optimizer with a base learning rate of $2 \times 10^{-5}$ for LLMs and $2 \times 10^{-4}$ for other parameters across all backbone VLMs. We also use a linear warmup scheduler for 4 epochs on the \datasetname. Models are trained for 30 epochs with a batch size of 8. After confirming that freezing LLMs leads to performance degradation, we unfreeze the LLMs during training. Additionally, we freeze all visual encoders during training.

\end{document}